\documentclass[a4paper,twoside]{article}

\usepackage{comment}
\usepackage{epsfig}
\usepackage{subcaption}
\usepackage{calc}
\usepackage{amssymb}
\usepackage{amstext}
\usepackage{amsmath}
\usepackage{amsthm}
\usepackage{multicol}
\usepackage{pslatex}
\usepackage{apalike}
\usepackage{pifont}
\usepackage{booktabs}
\usepackage{SCITEPRESS}
\usepackage{xcolor}
\newcommand{\unet}{U-Net}
\newcommand{\msgnet}{MSG-Net}
\newcommand{\umsgnet}{U-MSG-Net}
\newcommand{\gdsr}{GDSR}
\newcommand{\mpisintel}{MPI Sintel}
\newcommand{\middlebury}{Middlebury 2014}

\begin{document}

\title{Can We Use Neural Regularization to Solve Depth Super-Resolution?}

\author{\authorname{Milena Gazdieva\sup{1}\orcidAuthor{0000-0003-0047-1577}, Oleg Voynov \sup{1}\orcidAuthor{0000-0002-3666-9166}, Alexey Artemov\sup{1}\orcidAuthor{0000-0001-5451-7492}, Youyi Zheng\sup{2}\orcidAuthor{0000-0002-9120-9592}, Luiz Velho\sup{3}\orcidAuthor{0000-0001-5489-4909} and Evgeny Burnaev\sup{1}\orcidAuthor{0000-0001-8424-0690}}
\affiliation{\sup{1}Skolkovo Institute of Science and Technology, Moscow, Russia}
\affiliation{\sup{2}State Key Lab, Zhejiang University, Hangzhou, China}
\affiliation{\sup{3}Instituto Nacional de Matemática Pura e Aplicada, Rio de Janeiro, Brazil}
\email{\{milena.gazdieva, oleg.voynov, a.artemov\}@skoltech.ru, youyizheng@zju.edu.cn,  lvelho@impa.br, e.burnaev@skoltech.ru}
}


\abstract{Depth maps captured with commodity sensors often require super-resolution to be used in applications. In this work we study a super-resolution approach based on a variational problem statement with Tikhonov regularization where the regularizer is parametrized with a deep neural network. This approach was previously applied successfully in photoacoustic tomography. We experimentally show that its application to depth map super-resolution is difficult, and provide suggestions about the reasons for that.}

\onecolumn \maketitle \normalsize \setcounter{footnote}{0} \vfill

\section{\uppercase{Introduction}}
\label{sec:introduction}
Existing research reveals two major classes of state-of-the-art approaches to depth super-resolution (SR):  data-driven methods based on deep neural networks~\cite{pdn,msgnet,voynov} and variational optimization-based approaches~\cite{srfs2,srfs1}. 
Among these, learning-based methods bring the promise of leveraging powerful data-driven priors by learning these directly from data, which has proven to achieve impressive quantitative performance; however, as deep networks are trained by optimizing their target functions in an averaged sense, they are likely to produce imperfect estimates for specific (unseen) test instances.
In contrast, variational approaches commonly employ sophisticated hand-crafted regularizers and come with theoretical convergence guarantees, bounding an error between an estimate and the true high-resolution depth for individual instances.
Unfortunately, in some instances, the designed regularizer fails to capture image variations present in the real-world data, leading to suboptimal performance during variational optimization.

Combining advantages offered by these classes of approaches represents a natural interest; one particular variant is to incorporate an informative prior or a distribution statistic learned from a representative dataset of high-resolution depth images directly into the optimization formulation, e.g., as a learned regularizer.
In this approach, a pre-trained neural network aims to provide a data-driven loss term during the optimization of each individual test instance; we use this approach in this work.
The key intuition of it is to help alleviate the limitations of either learning- or optimization-based methods by leveraging the data distribution information summarized by pre-training on high-resolution depth images but still optimizing for each individual test instance.

Image-based tasks have successfully integrated data-driven loss functions~\cite{gatys2016neural,johnson2016perceptual}; inspired by this line of research, in this paper, we consider the question: ``Can we use learned regularizers to improve \textit{depth image super-resolution?}''
More specifically, we opted to use Network Tikhonov (NETT)~\cite{nett} framework, a theoretically sound regularization approach in the functional analytic sense offering attractive convergence properties, but adapt it to target specifically depth images.
To this end, we focus on training a deep convolutional neural network (CNN) to penalize artifacts found in an input depth image; we further integrate the norm of feature activations computed in a pre-trained network as a regularizer into an optimization procedure to operate on a per-image basis.
To the best of our knowledge, we are the first to investigate data-driven regularizers in the context of depth SR; we denote this approach DSRNETT.

We systematically study both theoretical requirements and empirical behavior of DSRNETT, where we are able to meet all theoretical claims (e.g., coercivity), obtain pre-trained regularization models, and perform optimization using a pre-trained network.
However, we discover that it is extremely difficult for the DSRNETT optimization procedure to reach a good solution; moreover, the respective target function is found to correlate weakly with commonly used quality measures such as per-pixel RMSE and perceptually-based RMSE$_v$~\cite{voynov}.
These results hold across multiple appropriate data augmentation strategies, optimization formulations, and for networks of varying capacity; together, we believe they indicate limitations of the NETT-type frameworks for single-image depth SR tasks.

\section{\uppercase{Related work}}
\label{sec:related}

\noindent \textbf{Depth super-resolution. }
Numerous approaches to deal with depth super-resolution problem were proposed in recent years. One group of approaches focuses on application of convolutional neural networks (CNNs) to depth SR (e.g., \cite{msgnet,voynov,iterresid}). For instance, \cite{msgnet} proposes a neural architecture that complements low-resolution depth features with high-frequency features from high-resolution RGB data, using a multi-scale fusion strategy. \cite{voynov} focus on designing a visual-difference based loss function, aiming to improve the performance of existing state-of-the-art depth processing methods. \cite{iterresid} proposes iterative residual learning based framework with the use of channel attention, multi-stage fusion and weight sharing strategies to tackle both synthetic and real-world degradation processes of depth maps.

Variational approaches represent another group, dealing with designing and optimizing an appropriate target function without relying on learning.  \cite{srfs2} propose a variational functional to jointly solve single-shot depth SR and shape-from-shading problems, i.e., inferring high-resolution depth from color variations in the high-resolution RGB image. 
\cite{srfs1} modifies the same approach for multi-shot depth SR using photometric stereo.

We focus on a combined approach; in contrast to variational approaches, it does not require to construct a regularizer manually but inherits their good convergence properties.
Unlike purely learning-based methods, it does not rely on training only but optimizes the solution for individual samples.

\noindent \textbf{Combined approaches} propose learning a regularizer and have been drawing attention recently; these have not been previously applied to depth SR.

One type of combined approaches corresponds to bilevel optimization, which incorporates learning into the definition of optimal regularizer parameters (e.g., \cite{bileveltv,bilevelmrf2}). According to this method, optimal parameters are derived as empirical risk minimizers for supervised training dataset. In \cite{bileveltv} bilevel optimization approach is applied for total variation (TV) type regularizers. Bilevel optimization with richer markov random field (MRF) parametrization of regularizer is investigated in \cite{bilevelmrf2}  by applying it to the image restoration tasks.

Data-driven loss functions have been successfully applied in the context of image-based tasks such as style transfer~\cite{gatys2016neural} and photo super-resolution~\cite{johnson2016perceptual} (where they are known as \textit{perceptual losses}), 
reconstruction tasks in computed and photoacoustic tomography~\cite{nett,augnett,datadriven,adversarial},
or for feature-based knowledge distillation (see, e.g.,~\cite{gou2021knowledge} for a survey).

Two recent combined approaches \cite{adversarial,nett} focus on neural network parametrizations of regularizers in application to computed and photoacoustic tomography reconstruction respectively.
\cite{adversarial} train a regularizer to discriminate ground-truth and (known) imperfect solutions.
Another combined approach \cite{nett} trains a regularizer to penalize solutions with various kinds of artifacts. 
\cite{datadriven} gives an overview of more methods aiming to incorporate learning into the variational optimization.
In this work, we parametrize the regularizer using a neural network;  more specifically, we follow~\cite{nett} that focuses on theoretical foundations, provides regularizing properties of their method, and derives respective convergence and convergence rates.

\section{\uppercase{Network Tikhonov Variational Formulation for Depth Super-Resolution}}
\label{sec:nett_theory}

We start with a brief review of the NETT formulation, rewriting the mathematical relations for our task. 
The goal of depth super-resolution is to 
\begin{equation}
    \label{eq:depth_sr_formulation}
    \text{estimate } x \in D\text{ from data } y_{\delta} = F(x)+\varepsilon_{\delta},
\end{equation}
where $F: D \subset X\rightarrow Y$ is a (known) downsampling operator, $X$, $Y$ denote the spaces of high- and low-resolution depth images, respectively, and $\varepsilon_{\delta}$ is an unknown data error, s.t., $\|\varepsilon_{\delta}\|\leqslant\delta$, $\delta>0$.

Depth super-resolution as formulated by~\eqref{eq:depth_sr_formulation} is an ill-posed inverse problem and admits many possible solutions.
To favor a particular class of solutions, Tikhonov regularization rewrites the task in an optimization perspective:
\begin{equation}
\label{eq:tiknonov_general_functional}
\mathcal{D}(F(x),y_{\delta})+\alpha\mathcal{R}(x)\rightarrow \min_{x\in D},
\end{equation}
where $\mathcal{D}: Y \times Y \rightarrow [0,\infty)$ is a data fidelity term defined on depth images and $\mathcal{R}: D \rightarrow[0,\infty)$ is a regularization term.
In this work, we focus on the form of regularizer term, seeking to learn it from data rather than design it manually.
Our DSRNETT draws inspiration from Network Tikhonov framework~\cite{nett}, but is redesigned to target specifically depth images.
Let $\Phi$ denote a neural network, which we view as a composition of affine linear maps $\mathbb{V}_l(x)=\mathbb{A}_l x + b_l$ and non-linearities $\sigma_l$ where $l$ is a layer index.
Many convolutional architectures such as \unet~\cite{unet} admit this form.
We indicate a (trainable) affine part $\mathbb{V} = (\mathbb{V})_{l}$ in the notion of $\Phi_{\mathbb{V}}(\cdot)$ and formulate our regularizer as
\begin{equation}
\label{eq:our_nett_regularizer}
\mathcal{R}(x) = \psi(\Phi_{\mathbb{V}}(x)),
\end{equation}
where $\psi:X\rightarrow [0,\infty)$ is a scalar function.

\begin{figure*}[t!]

\begin{center}
\includegraphics[width=0.995\textwidth]{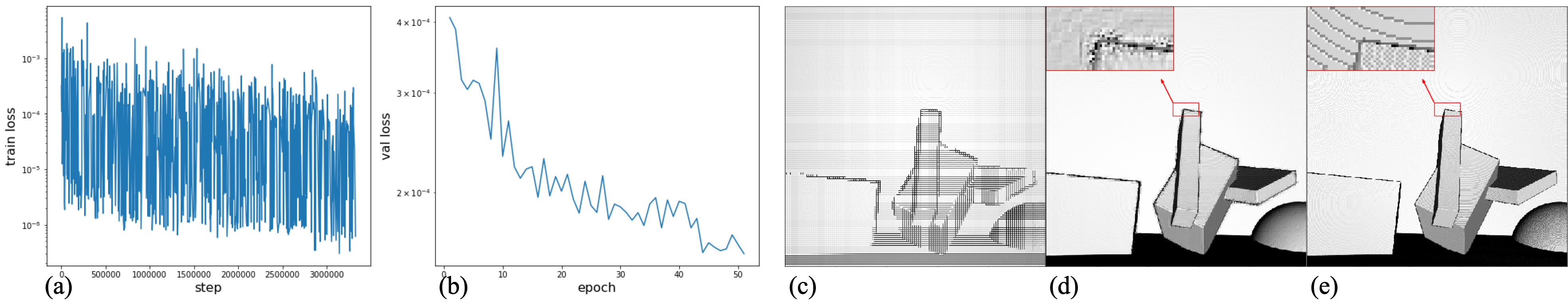}
\caption{Typical pre-training results using \gdsr:
(a): train loss dynamics; 
(b): validation loss dynamics, log-scale; 
(c): initial approximation (network input); 
(d): reconstructed depth image; 
(e): ground-truth.
} 
\label{fig:gdsr_exp_1_training}
\end{center}

\begin{center}
\includegraphics[width=0.995\textwidth]{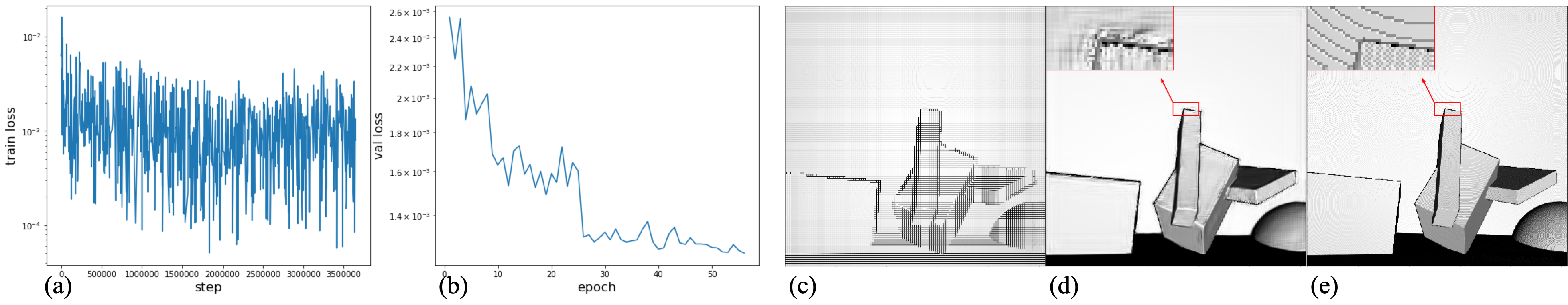}
\caption{Typical pre-training results using noisy \gdsr:
(a): train loss dynamics; 
(b): validation loss dynamics, log-scale; 
(c): initial approximation (network input); 
(d): reconstructed depth image; 
(e): ground-truth.} 
\label{fig:gdsr_exp_2_training}
\end{center}

\begin{center}
\includegraphics[width=0.825\textwidth]{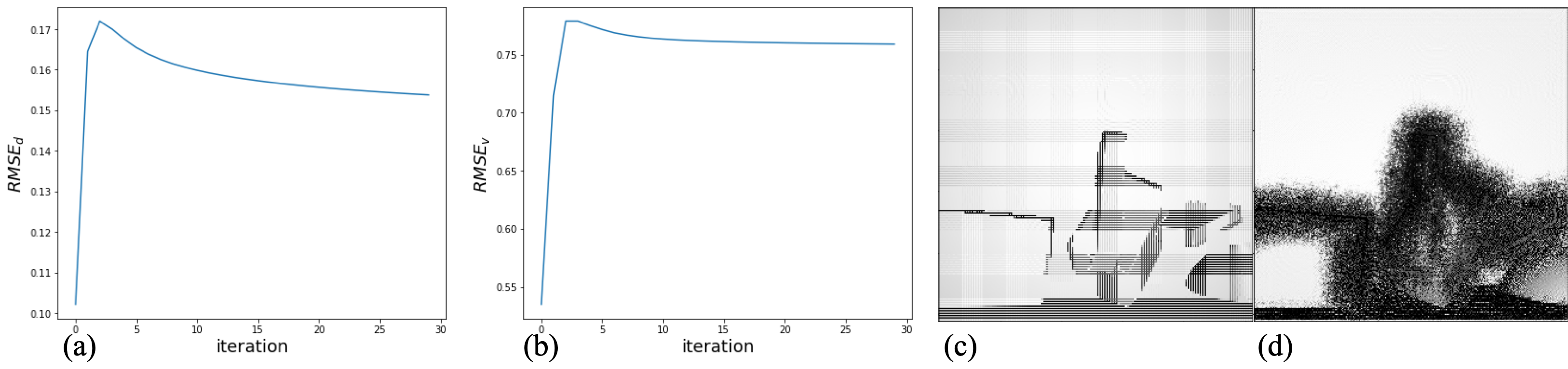}
\caption{Typical optimization results using networks pre-trained on \gdsr:
(a): RMSE$_d$ to ground-truth dynamics; 
(b): RMSE$_v$ to ground-truth dynamics;
(c): initial approximation;
(d): optimization result (30 iterations).} 
\label{fig:gdsr_exp_1_opt}
\end{center}
\begin{center}
\includegraphics[width=0.825\textwidth]{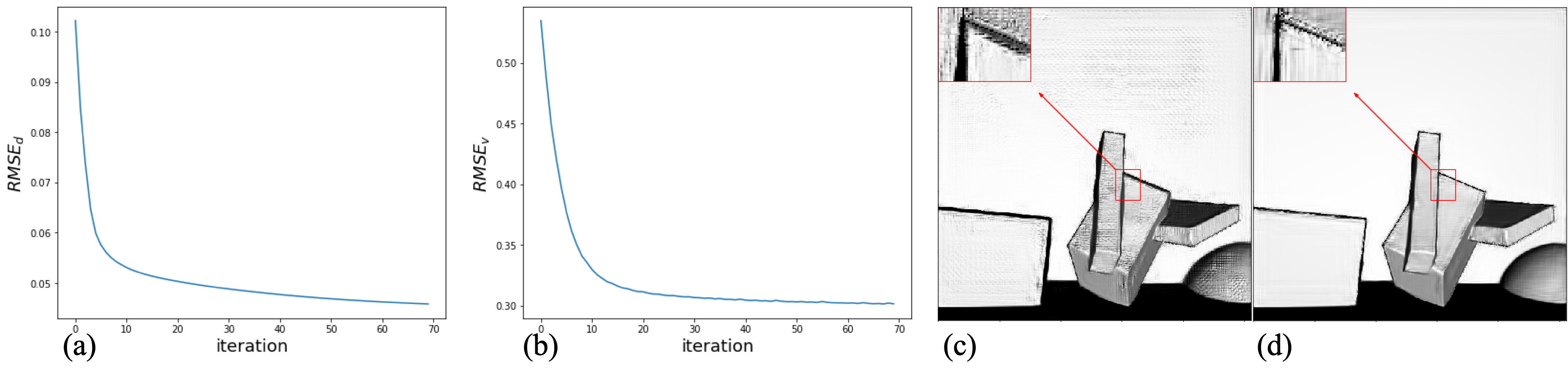}
\caption{Typical optimization results using networks pre-trained on noisy \gdsr:
(a): RMSE$_d$ to ground-truth dynamics; 
(b): RMSE$_v$ to ground-truth dynamics;
(c): initial approximation;
(d): optimization result (70 iterations).
} 
\label{fig:gdsr_exp_2_opt}
\end{center}

\end{figure*}

To effectively guide optimization in~\eqref{eq:tiknonov_general_functional}, network parameters $\mathbb{V}$ must be learned using the available training data before any optimization may happen (we assume that non-linearities $\sigma_l$ are not trainable).
We consider the two following pre-training schemes.

\textit{Scheme 1:} Train the network to predict an artifact component from the input depth image.
In this scenario, we split a set $X^{\text{HR}}$ of high-resolution training images $x_i$ into two disjoint subsets $X^{\text{HR}}_0 \cup X^{\text{HR}}_1$ and compute approximations $\widetilde{x}_i = F^+(F(x_i))$ for each $x_i \in X^{\text{HR}}_1$,
using an upsampling operator $F^+$ (a pseudo-inverse of $F$).
The network then trains to predict a residual $r_i = x_i - \widetilde{x}_i$ from the input approximation $\widetilde{x}_i$ for images in $X^{\text{HR}}_1$, or an exactly zero image for images in $X^{\text{HR}}_0$. 
A simple regularizer based on this network is a Euclidean norm of the predicted residual: $\mathcal{R}(x) = \|\Phi_{\mathbb{V}}(x)\|_2^2$.

\textit{Scheme 2:} An alternative approach, which we experimentally observed to aid convergence, is to train the model to directly predict ground-truth depth images given either an input approximation or a ground-truth image.
In this scenario, the network is trained to output the same image as input  $x_i$ for images in $X^{\text{HR}}_0$, and an ideal $x_i$ from input approximation $\widetilde{x}_i$ for images in $X^{\text{HR}}_1$.
We attribute the advantages of this approach to the benefit of having inputs of similar scale.
A similar approach is employed by \cite{augnett}.
We then keep the idea of penalizing images with artifacts in the design of regularizer, which is defined as
$\mathcal{R}(x) = 
\|\Phi_{\mathbb{V}}(x)-x\|_2^2$.

NETT is guaranteed to converge to a unique and stable solution provided that certain analytic conditions hold \cite{nett}.
We report a detailed analysis in the context of our task in the Appendix.

\section{\uppercase{Experiments}}
\label{sec:experiments}

\subsection{Toy Examples: GDSR and SimGeo}
\label{subsec:gdsr}

We start by implementing our approach on simple synthetic depth images, using the dataset from \cite{pdn}, that we denote GDSR.
It consists of scenes containing randomly placed cubes, spheres and planes in varying poses and dimensions. 
Here and below, we visualize 3D scenes as renderings of corresponding depth maps, obtaining these using a simple normals-based method~\cite{voynov}.

We train two CNNs to predict residuals in upscaled depth images according to \textit{Scheme 1} (i.e., the CNN predicts the artifact part of the training instances): one using the original depth images and another using the noise-augmented depth images, and perform optimization using the two trained models as regularizers.
To optimize for a final depth image, we minimize a sum of the squared $L_2$-norm data fidelity term and the squared $L_2$-norm of network output:
\begin{equation}
\frac{1}{2}\|F(x) - y\|_2^2 + \alpha \|\Phi_{\mathbb{V}}(x)\|_2^2 \rightarrow \min_{x}.
\label{eq:loss_gdsr_exps}
\end{equation}

Since the network is trained to output artifact part of the depth maps for visualization of its performance we use rendering of sum of network input and network output, which should approximate ground-truth depth map.

While we are able to train efficient estimators of residuals using both clean and noise-augmented images, as shown in Figures~\ref{fig:gdsr_exp_1_training}--\ref{fig:gdsr_exp_2_training}, using the noise augmentations proves to have a major effect on the optimization performance (see Figures~\ref{fig:gdsr_exp_1_opt}--\ref{fig:gdsr_exp_2_opt}).  
More specifically, we find that regularizing optimization in~\eqref{eq:loss_gdsr_exps} with the network trained on clean synthetic images leads to a bad solution (see Figures~\ref{fig:gdsr_exp_1_opt} (c)--(d)).

\begin{figure}[h!]
\begin{center}
\includegraphics[width=\columnwidth]{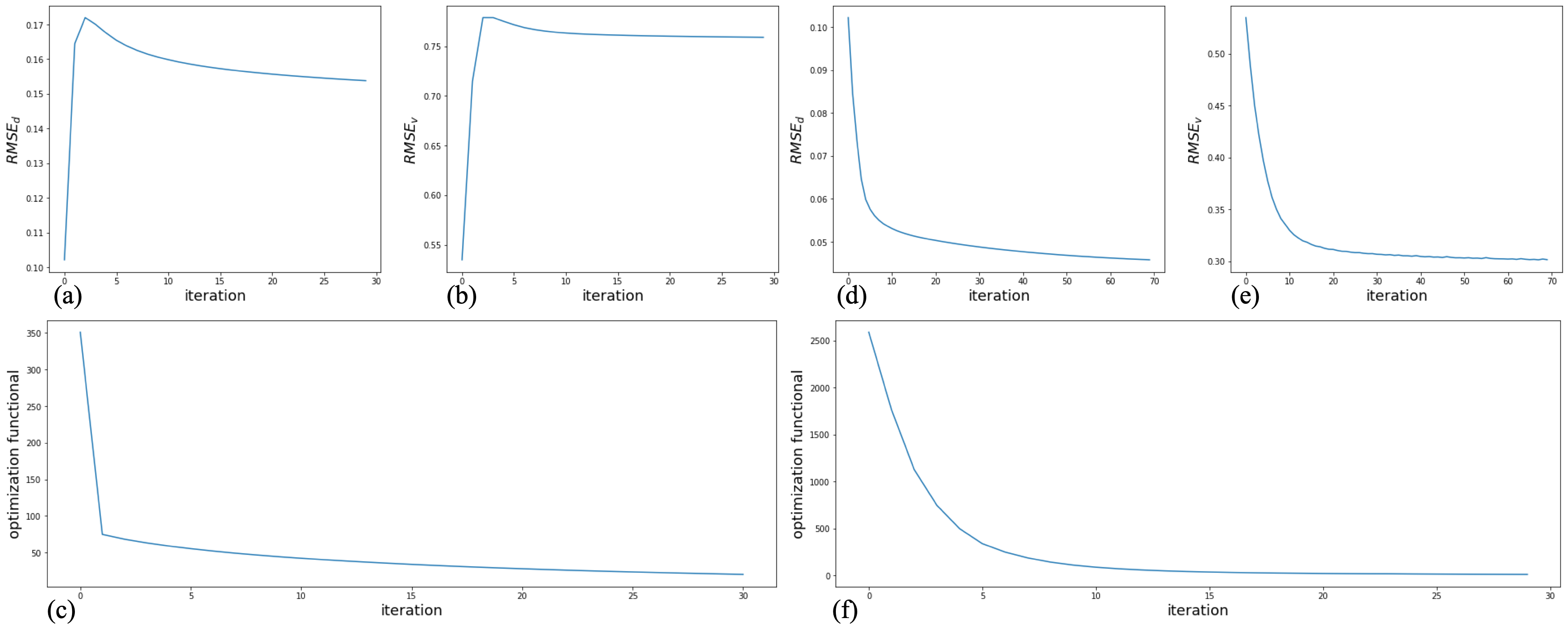}
\caption{Pre-training on clean data (a)--(c) leads to optimization results where the DSRNETT loss function~\eqref{eq:loss_gdsr_exps} does not correlate well with the commonly used RMSE$_d$ and RMSE$_v$ measures.
Pre-training with noise-augmented data (d)--(f) yields more predictable optimization dynamics.}
\label{fig:gdsr_exp_1_rmse_vs_functional}
\end{center}
\end{figure}

Moreover, we discover that the DSRNETT loss function correlates poorly with RMSE$_d$, the standard root-mean-squared (RMS) error capturing per-pixel differences between depth maps, and perceptual measure RMSE$_v$~\cite{voynov}, defined as the RMS difference between renderings capturing visual differences between 3D surfaces represented as depth maps.
Both RMSE$_d$ and RMSE$_v$ represent measures commonly used in evaluating depth map processing performance (see inset Figure~\ref{fig:gdsr_exp_1_rmse_vs_functional}).

Eventually, optimization resulted in only a marginal improvement for validation samples from GDSR (see Figures~\ref{fig:gdsr_exp_2_opt} (c)--(d)).
We hypothesize that this stems from a limited capacity in our network, and seek to increase it in Section~\ref{subsec:middlebury}.

\noindent \textbf{Implementation details.}
We obtain $128\times 128$ high-resolution training samples by splitting each image from \gdsr~ dataset into 16 patches.
For 50\% of these high-resolution patches, we generate their low-resolution counterparts, choosing $F$ as a box downsampling method with a scaling factor of 4, and obtain approximations $\widetilde{x}$ by applying the pseudo-inverse $F^+$ as an upsampling operator.
In total, we obtained 128K patches for training and 32K for validation.

We employ a \unet-like architecture~\cite{unet}. 
In order to satisfy the coercivity condition  for the regularizer term, which is stated by \cite{nett} to be the most restrictive one, we replace ReLU activation function in network architecture with leaky ReLU.
We train by optimizing MSE loss using Adam~\cite{kingma2015adam} for ${\sim}70$ epochs using a batch size of 2.

We perform optimization in~\eqref{eq:loss_gdsr_exps} using the incremental gradient descent algorithm following~\cite{nett}. 
Thus, optimization consisted of two steps: gradient descent step for data fidelity term with weight $s$ and gradient descent step for regularizer term with weight $s \cdot \alpha$, done with the use of backpropagation algorithm. 
We also choose the step weights, following \cite{nett}. Yet in contrast, we do not use a zero image as initial approximation, but start optimization from approximation of ground-truth depth map, derived by upsampling $y$, since the network is trained on similar train samples.

To train the network with noisy data, we add Gaussian noise with variance equal to $0.05$ to all pixels in input depth images.

\begin{figure}[t!]
\centering
\includegraphics[width=\columnwidth]{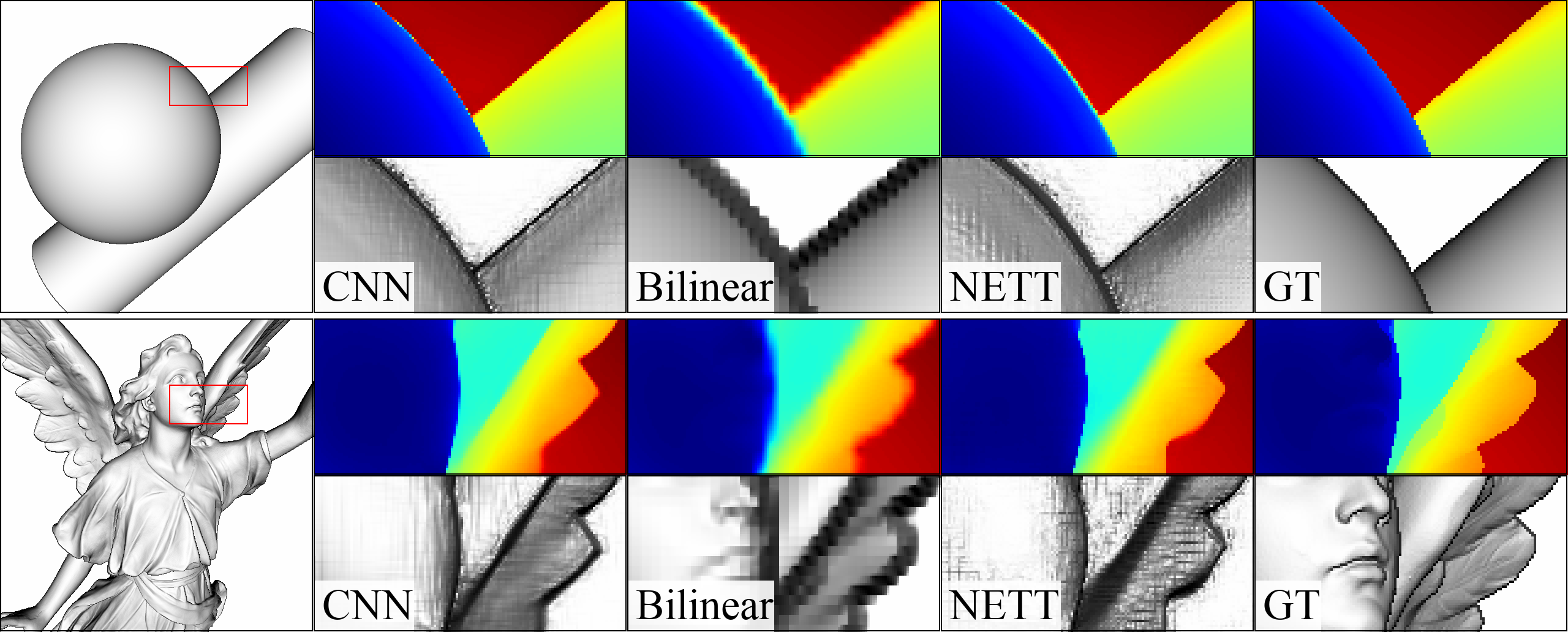}
\caption{Performance of NETT approach in comparison to underlying network (CNN) and Bilinear interpolation on samples from SimGeo for a network pre-trained on \gdsr~ with added noise. For each sample - First row: depth maps in pseudo-colour; Second row: renderings in grayscale.}
\label{fig:toy_nett}
\end{figure}

Due to the low complexity of the training dataset, network was pre-trained very well and optimization did not improve network result for validation samples from \gdsr. 
On the other hand, for geometrically more complex synthetic scenes from SimGeo~\cite{voynov} dataset network, pre-trained on \gdsr, output poor result and NETT approach was not working correctly; see Figure \ref{fig:toy_nett} for comparison of approach performance with underlying network and bilinear interpolation on simple and complex synthetic scenes. According to these suggestions we decided to reimplement NETT with stronger network architecture, using complex synthetic and real-world scenes for pre-training.

\subsection{Complex Synthetic and Real-world Scenes: \mpisintel~ and Middlebury}
\label{subsec:middlebury}

\begin{figure*}[t!]

\begin{center}
\includegraphics[width=0.995\textwidth]{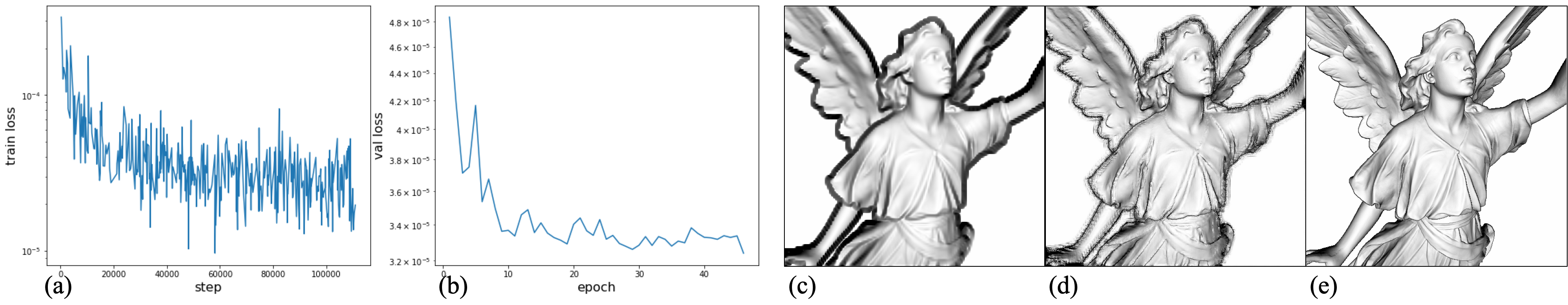}
\caption{Typical pre-training results using \middlebury~/ \mpisintel~(test sample from SimGeo):
(a): train loss dynamics; 
(b): validation loss dynamics, log-scale; 
(c): input approximation (network input); 
(d): reconstructed depth image; 
(e): ground-truth image.} 
\label{fig:middlebury_exp_1_training}
\end{center}

\begin{center}
\includegraphics[width=0.825\textwidth]{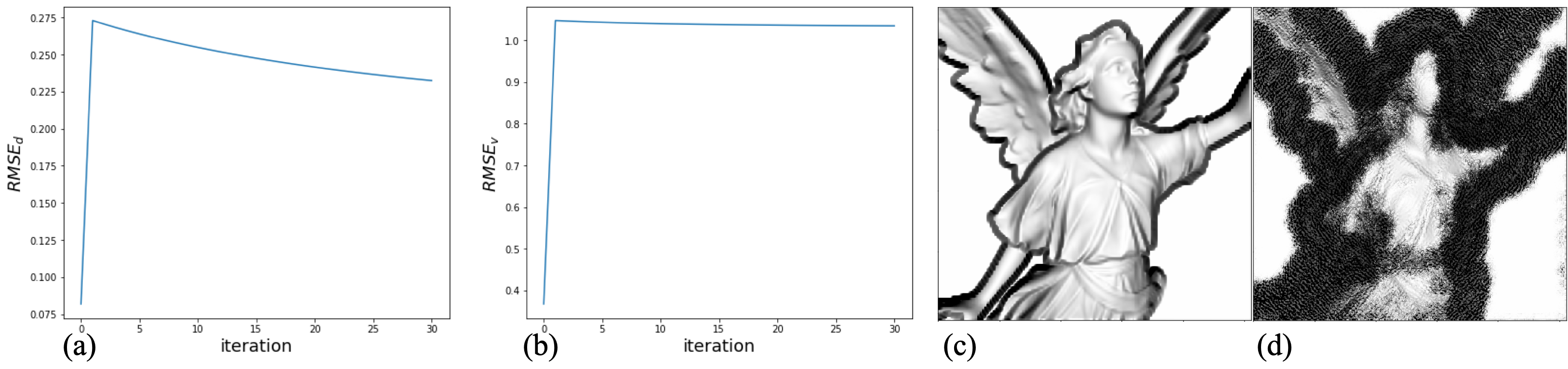}
\caption{Typical optimization results using networks pre-trained on \middlebury~ and \mpisintel:
(a): RMSE$_d$ to ground-truth dynamics; 
(b): RMSE$_v$ to ground-truth dynamics;
(c): initial approximation;
(d): optimization result (30 iterations).} 
\label{fig:middlebury_exp_1_opt}
\end{center}

\begin{center}
\includegraphics[width=0.825\textwidth]{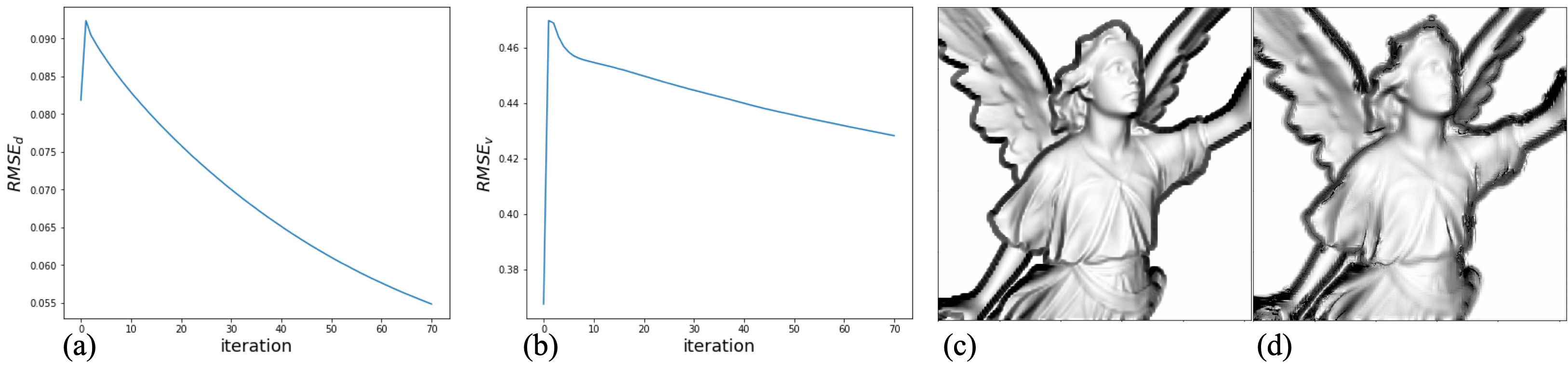}
\caption{Typical optimization results using networks pre-trained on noisy \middlebury~ and \mpisintel:
(a): RMSE$_d$ to ground-truth dynamics; 
(b): RMSE$_v$ to ground-truth dynamics;
(c): initial approximation;
(d): optimization result (70 iterations).} 
\label{fig:middlebury_noise_opt}
\end{center}

\begin{center}
\includegraphics[width=0.825\textwidth]{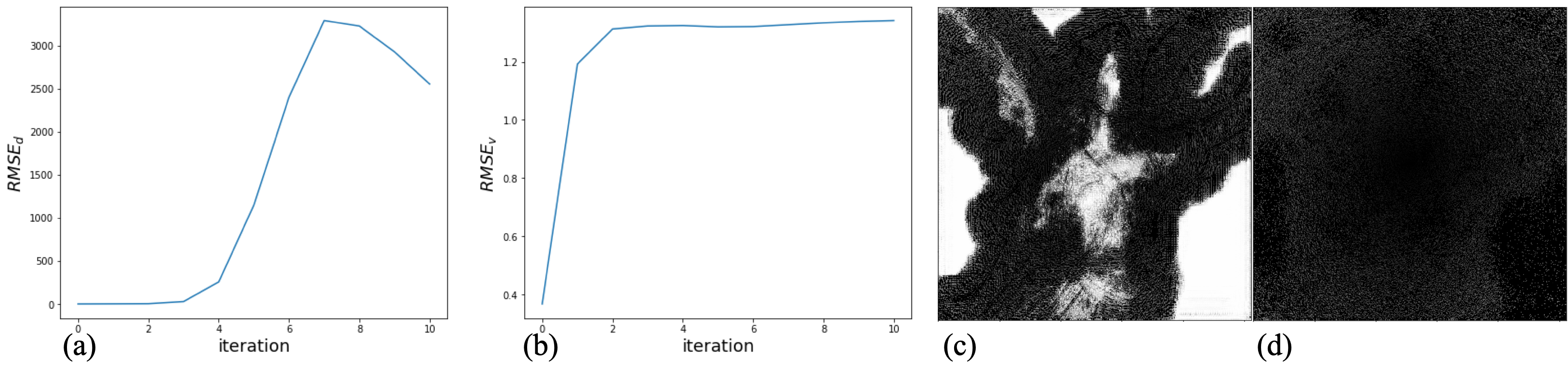}
\caption{Typical optimization results using networks pre-trained on \middlebury~ and \mpisintel~ augmented using linear interpolation between GT and approximation:
(a): RMSE$_d$ to ground-truth dynamics; 
(b): RMSE$_v$ to ground-truth dynamics;
(c): initial approximation;
(d): optimization result (7 iterations).} 
\label{fig:middlebury_approach1_opt}
\end{center}

\end{figure*}

To study our approach in a full-scale setting, we use data collections and network architectures with larger complexity. 
More specifically, we use \umsgnet, a \unet-like version of the MSG-Net architecture~\cite{msgnet} that has demonstrated good performance for depth SR due to the effective use of RGB guidance (see Appendix for more details).
For training our network, we use \mpisintel~\cite{mpisintel} and Middlebury 2014~\cite{middlebury} datasets, both providing RGB-D images.
\mpisintel~ contains complex scenes retrieved from a naturalistic 3D animated short film, while Middlebury 2014 consists of complex high-quality real images captured with a structured light system.

Due to the difficulties with optimization convergence, that we discuss further, in addition to training the network according to \textit{Scheme 1} we experimented with training according to \textit{Scheme 2}, where the goal is to predict a reconstructed image given an input approximation. We observed that \textit{Scheme 2} leads to better convergence in comparison to \textit{Scheme 1} and further provide results for \textit{Scheme 2}.
For optimization, we accordingly set the regularizer to $\mathcal{R}(x)=\|\Phi_{\mathbb{V}}(x) - x\|^2_2$, re-using the general framework described previously.

We do not explicitly account for the coercivity condition of the architecture, since \cite{nett} provides ways to obtain coercivity for an arbitrary architecture, e.g., using skip connection between network input and output, as we discuss further.

\noindent \textbf{Data augmentation.}
To aid generalization, we study a number of data augmentation strategies aiming to expand the training domain, summarized in Table~\ref{tab:all_experiments_results}.

\textit{Training with noisy inputs.}
We train the network on input depth images augmented with different variations of additive random noise (lines 2, 4-6 in the table).

\textit{Training with both noisy inputs and targets.}
Aiming to increase network robustness, we train the network to predict noisy ground-truth depth from noisy ground-truth, keeping target noise variance smaller than that of input noise (line 7).

\textit{Training with input in-between approximation and ground truth.}
The regularizer $\mathcal{R}(x)$ is designed to push $x$ to ground-truth data manifold, approximated by $\Phi_{\mathbb{V}}(x)$.
Based on the derivation in Appendix, we hypothesize that, in contrast to this desired behaviour,
the optimization of the regularizer also pushes the output of the network in the undesired direction.
To prevent that, we additionally train the network on the samples,
where the input is the result of one step of optimization with small regularizer step size $s\cdot\alpha=0.001$.
We also try a simpler strategy to train on random linear interpolation between the approximation and the ground truth as the input.

Additionally, in the experiments with \mpisintel~ and \middlebury~ we randomly rotate the patches by $90^{\circ}$.
We briefly review the results of these experiments below.

\begin{table*}[t]
    \centering
    \caption{Summary of conducted experiments. Mid./Sint. correspond to \middlebury/\mpisintel.}
    \resizebox{\textwidth}{!}{
    \begin{tabular}{@{}lllllllll@{}}
        \toprule
        Pre-training & Net       & Scheme & Input augmentation  & Optimization & Result        & Opt.       & \(\mathrm{RMSE_d}\)     & \(\mathrm{RMSE_v}\)    \\ \midrule
        GDSR         & U-Net     & 1      & ---                                                                                                                                                                   & GDSR         & Noisy         & \checkmark           &  \checkmark          & \checkmark          \\
        GDSR         & U-Net     & 1      & Gaussian noise with $\sigma = 0.05$                                                                                                                                           & GDSR         & OK            &      \checkmark      &     \checkmark       & \checkmark          \\ \midrule
        Mid./Sint.   & U-MSG-Net & 2      & ---                                                                                                                                                                   & SimGeo       & Noisy         &  \checkmark          &   \ding{55}         &    \ding{55}       \\
        Mid./Sint.   & U-MSG-Net & 2      & $\sigma = 0.03$                                                                                                                                           & SimGeo       & Over-smoothed &  \checkmark          &   \checkmark         &   \checkmark        \\
        Mid./Sint.   & U-MSG-Net & 2      & $\sigma = 0.03$, scaled by 0.7 every 10-th epoch                                                   & SimGeo       & Over-smoothed &  \checkmark          &  \checkmark          & \checkmark          \\
        Mid./Sint.   & U-MSG-Net & 2      & Additional input samples (+): GT with gaussian noise with $\sigma = \mathrm{MSE(GT, approximation)}$                                                              & SimGeo       & Same as input & \checkmark & \checkmark & \ding{55} \\
        Mid./Sint.   & U-MSG-Net & 2      & + GT with noise $\epsilon$, and target with noise $\epsilon / 10$                      & SimGeo       & Same as input & \checkmark & \checkmark & \ding{55} \\
        Mid./Sint.   & U-MSG-Net & 2      & + random linear interpolation between GT and approximation & SimGeo       & Noisy &  \ding{55}          &    \ding{55}        &      \ding{55}     \\
        Mid./Sint.   & U-MSG-Net & 2      & + with the result after one step of optimization                                          & SimGeo       &   Same as input       & $\checkmark$ & \ding{55}  & \ding{55} \\
        \bottomrule
    \end{tabular}
    }
    \label{tab:all_experiments_results}
\end{table*}

Similarly to results in Section~\ref{subsec:gdsr}, training without data augmentations results in optimization hitting a bad solution (see Figure~\ref{fig:middlebury_exp_1_opt}).
Network trained on noisy data tends to produce over-smooth depth maps (see Figure~\ref{fig:middlebury_noise_opt}).
Using noisy data with gradually decreasing noise variance similarly leads to over-smoothed network outputs and optimization result.
Expanding training data with intermediate optimization results did not alleviate difficulties with optimization  convergence.
Moreover, in some instances, this resulted in well trained network, but exploding optimization, yielding extremely noisy results (see Figure \ref{fig:middlebury_approach1_opt}).
Training using noisy targets again led to over-smoothed network outputs and aggravated artefacts in optimization  results.

\noindent \textbf{Implementation details.}
We construct training dataset similar to the one proposed in \cite{msgnet} for training \msgnet~ but appropriate for use within DSRNETT.
We split all images into patches of size $64\times64$. 
For all patches we calculate intensity, corresponding to RGB component.
For half of the patches we generate low-resolution depth using box downsampling method with a scaling factor of 4, and obtain approximations using a bilinear upsampling method instead of pseudo-inverse of box downsampling, since the former produces patches that are more relevant for training \msgnet. Finally, following our Scheme 2, we train the model using the high-frequency (HF) components of intensity and depth of generated patches as inputs, and HF components of ground-truth depth and intensity as targets. 
For simplicity we further omit the fact that model is trained on HF components and its inputs and outputs contain not only depth, but also intensity part, since it is not changed during training or optimization.
In total, training was performed on $75$K patches with $15$K patches used for validation. We train by optimizing MSE loss using Adam for ${\sim}120$ epochs with a batch size of 128.

We experimentally set minimum noise variance to $0.03$.
In experiment with noisy targets for network inputs we added Gaussian noise with variance  $\sigma = 0.001$; each target variance of noise was defined as $\text{MSE}(\sigma, 0)$, divided by $10$.

\noindent \textbf{Ensuring coercivity of regularizer.} 
According to remark in \cite{augnett} in order to ensure coercivity of regularizer in experiments with \umsgnet,  we employed skip-connection between network input and output and define regularizer as $\mathcal{R}(x) = \|\Phi_{\mathbb{V}}(x) - x\|^2 + \|x\|^2.$ We have checked, that such regularizer design does not help to solve optimization issues and concluded, that optimization convergence to undesired local minimum was not lead by possible regularizer non-coercivity.

\section{\uppercase{Discussion}}
\label{sec:discussion}
In this work, we performed an adaptation of NETT~\cite{nett}, an approach to image processing leveraging data-driven regularizers, for the depth super-resolution (SR) task.
We have validated that our formulation of depth SR meets all theoretical requirements of NETT, including the restrictive coercivity condition.
Furthermore, we were able to train an efficient residual estimator deep network in all experimental cases. 
Unexpectedly, we have discovered subsequent optimization to converge to the bad minimum in most experiments despite multiple efforts to increase network stability by varying the training dataset, applying data augmentations, selecting different models, or considering various formulations of the optimization task.

Our results raise questions regarding the use of learned regularizers in the context of depth SR (or image-based tasks), which may represent promising future research directions. 
More specifically, 
\begin{enumerate}
\item What is the required form of the regularization term that would allow effective optimization?

\item What are the characteristic features of image-based tasks that might prevent optimization methods such as \cite{nett} from converging to good solutions? 

\item What is the ``right'' training procedure for the regularizer?
\end{enumerate}

\section*{\uppercase{Acknowledgements}}

This work was supported by Ministry of Science and Higher Education grant No. 075-10-2021-068.
We acknowledge the usage of Skoltech CDISE supercomputer Zhores~\cite{zacharov2019zhores} for obtaining the presented results.

\bibliographystyle{apalike}
{\small
\bibliography{bib}}

\section*{\uppercase{Appendix}}
\subsection*{Network Tikhonov Regularization Assumptions}
\cite{nett} prove well-posedness and convergence of NETT regularization, provided that certain assumptions on regularizer and data fidelity term hold. 

We have tried to ensure fulfillment of this conditions in our work. Conditions on data fidelity term are not restrictive and hold for squared $L_2$-norm distance, while the conditions on regularizer term include:
\begin{enumerate}
    \label{eq:conditions}
    \item Regularizer $\mathcal{R}(x) = \psi(\Phi_{\mathbb{V}}(x))$ is weakly lower semicontinuous, which is guaranteed by conditions:
    \begin{itemize}
        \item Linear operators $\mathbb{A}_l$ are bounded;
        \item Non-linearities $\sigma_l$ are weakly continuous;
        \item Scalar functional $\psi$ is weakly lower-semicontinuous;
    \end{itemize}
    \item $\mathcal{R}(x)$ is coercive, i.e. $\lim_{\|x\|\rightarrow \infty}\mathcal{R}(x)= 0$.
\end{enumerate}

\unet-type and \umsgnet~ architectures include convolution operations and batch normalizations, that are bounded linear operations, and max pooling, leaky ReLU and upsampling operations, that are weakly continuous and coercive non-linearities.
We use squared $L_2$-norm distance as weakly lower semi-continuous scalar functional $\psi$. 
We note that the number of parameters in the \unet~ model is close to 7M, while in the \umsgnet~ it is around 700K.

Several ways to obtain coercivity of the regularizer are discussed in \cite{nett}. One way is to ensure layer-wise coercivity, i.e. use coercive non-linearities $\sigma_l$ and linear operators $\mathbb{A}_l$, satisfying the inequality:
\begin{equation}
    \exists c_l\in [0,\infty)\forall x\in X:\;\|x\|\leq c_l \|A_l x\|.
\end{equation}
We relied on this approach in experiments with \unet-type architecture.
Since \umsgnet~ contains non-coercive parametric ReLU non-linearities, we used another way to obtain network coercivity by exploiting skip connection between network input and output, see "Ensuring coercivity of regularizer" in Section~\ref{subsec:middlebury}. 

Specifically, fulfillment of presented conditions guarantee existence of a solution, stability and convergence for optimization problem \eqref{eq:tiknonov_general_functional}. Stability of the method corresponds to continuous dependence of the solutions $x$ of optimization problem \eqref{eq:tiknonov_general_functional} on input $y$. Convergence of the method states, that while noise level $\delta_k$ in inputs $y_{\delta_k}$ decreases to zero, sequence $x_k$ of corresponding solutions weakly converges to the $\mathcal{R}(\cdot)$-minimizing solution $x_+$ of $F(x)=y_0$, if it is unique.

\section*{Experiments}

\begin{figure}[t]
\centering
\includegraphics[width=\linewidth]{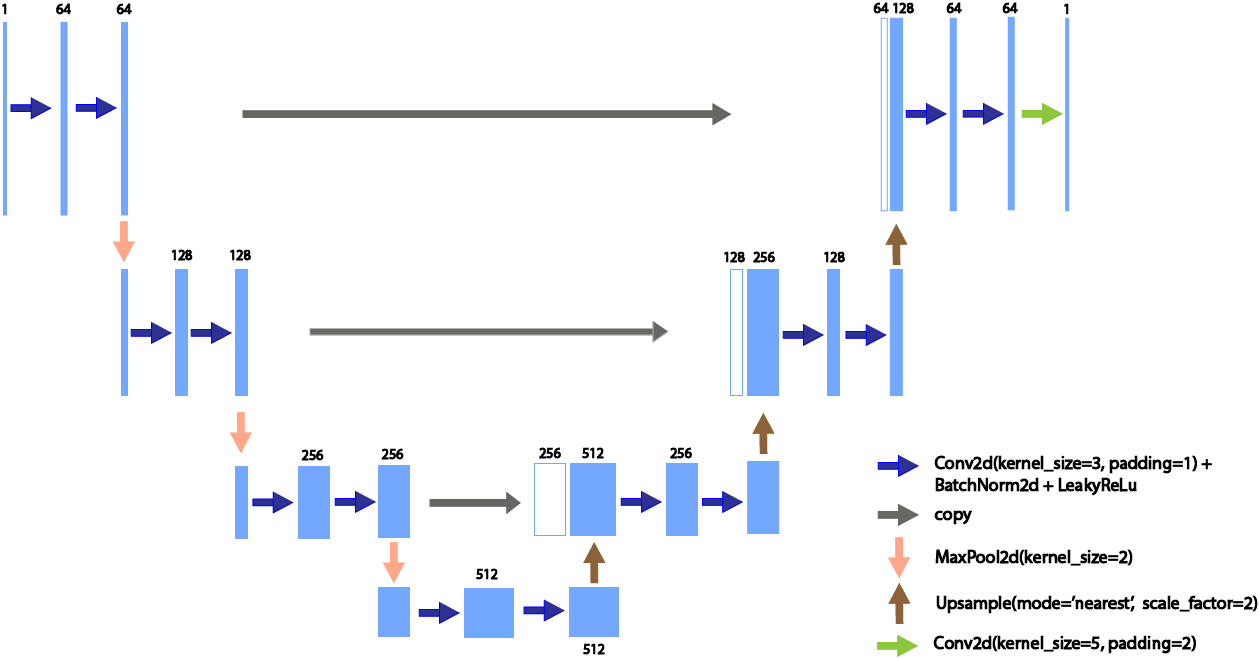}
\caption{\unet-type architecture.}
\label{fig:gdsr_unet}
\end{figure}

\begin{figure}[t]
\centering
\includegraphics[width=\linewidth]{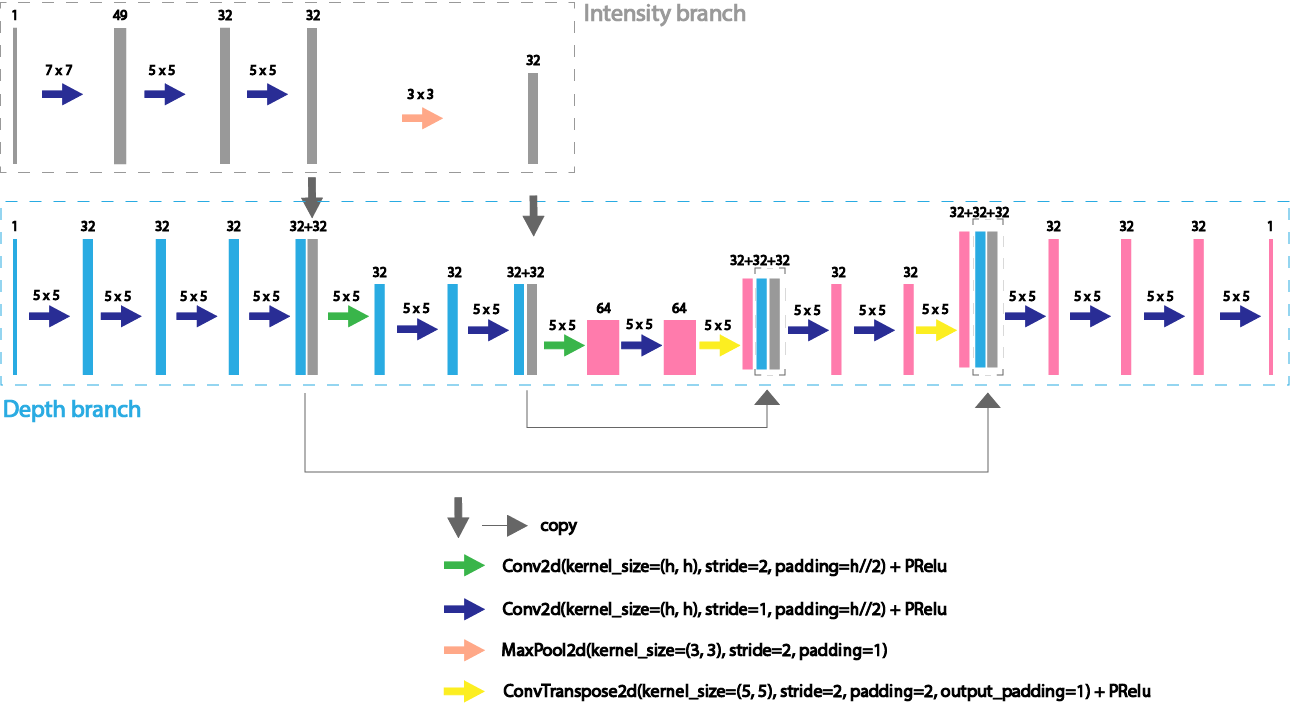}
\caption{\umsgnet~ architecture.}
\label{fig:middlebury_unetmsg}
\end{figure}

Figures \ref{fig:gdsr_unet} and \ref{fig:middlebury_unetmsg} represent architectures of networks, used in experiments in Sections \ref{subsec:gdsr} and \ref{subsec:middlebury} respectively.

\subsection*{Training on Intermediate Optimization Results as Input}
In our experiments we use approximation $\widetilde{x}$, i.e. upsampled low-resolution version of ground-truth depth, as optimization initial approximation. We suppose, that optimization approximation is not significantly changed after the first data fidelity gradient descent step and treat $\widetilde{x}$ as its result.

Then after the regularizer gradient descent step we obtain approximation $x_1$:
\begin{equation}
    x_1 = \widetilde{x} - s \cdot \alpha \cdot \frac{d \mathcal{R}(x)}{x}|_{x=\widetilde{x}}
\end{equation}
If network is trained to predict ground-truth depth given both $\widetilde{x}$ or $x_1$, then regularizer gradient descent step would not move optimization approximation to the direction of $x_1$:
\begin{align*}
    &\|\Phi_{\mathbb{V}}(x_1)-x_1\|^2 
    = \|\Phi_{\mathbb{V}}(\widetilde{x})-\widetilde{x} + s \cdot \alpha \cdot \frac{d \mathcal{R}(x)}{x}|_{x=\widetilde{x}}\|^2 \\
    &\geq \|\Phi_{\mathbb{V}}(\widetilde{x})-\widetilde{x}\|^2 + \|s \cdot \alpha \cdot \frac{d\mathcal{R}(x)}{x}|_{x=\widetilde{x}}\|^2\\
    &\geq \|\Phi_{\mathbb{V}}(\widetilde{x})-\widetilde{x}\|^2.
\end{align*}
Thus, we come to idea to add optimization first step approximation for small $s \cdot \alpha=0.001$ to the training samples.
\vfill
\end{document}